\documentclass{article}

    \PassOptionsToPackage{numbers, compress}{natbib}

\usepackage[preprint]{neurips_2021}

\usepackage[utf8]{inputenc} 
\usepackage[T1]{fontenc}    
\usepackage{url}            
\usepackage{booktabs}       
\usepackage{amsfonts}       
\usepackage{nicefrac}       
\usepackage{microtype}      
\usepackage{xcolor}         

\usepackage{caption}
\usepackage{subcaption}
\usepackage{lipsum}
\usepackage{graphicx}
\usepackage{flushend}

\graphicspath{{figures/}}

\title{Towards a Taxonomy of Graph Learning Datasets}

\author{
Renming Liu$^3$,
Semih Cantürk$^{1,2}$,\\
\textbf{Frederik Wenkel}$^{1,2}$,
\textbf{Dylan Sandfelder}$^{1,4}$,
\textbf{Devin Kreuzer}$^{1,4}$,
\textbf{Anna Little}$^5$,
\textbf{Sarah McGuire}$^3$, \\
\textbf{Leslie O'Bray}$^6$,
\textbf{Michael Perlmutter}$^7$,
\textbf{Bastian Rieck}$^8$, \\
\textbf{Matthew Hirn}$^3$,
\textbf{Guy Wolf}$^{1,2}$ and
\textbf{Ladislav Rampášek}$^{1,2}$
\\
\\
$^1$Mila - Quebec AI Institute,
$^2$Université de Montréal,
$^3$Michigan State University, \\
$^4$McGill University,
$^5$University of Utah,
$^6$ETH Zürich, \\
$^7$University of California, Los Angeles,
$^8$Helmholtz Zentrum München
\\
\texttt{mhirn@msu.edu, \{wolfguy, ladislav.rampasek\}@mila.quebec}
}

\begin{document}

\maketitle

\vspace{-15pt}
\begin{abstract}
Graph neural networks (GNNs) have attracted much attention due to their ability to leverage the intrinsic geometries of the underlying data. Although many different types of GNN models have been developed, with many benchmarking procedures to demonstrate the superiority of one GNN model over the others, there is a lack of systematic understanding of the underlying benchmarking datasets, and what aspects of the model are being tested.  Here, we provide a principled approach to taxonomize graph benchmarking datasets by carefully designing a collection of graph perturbations to probe the essential data characteristics that GNN models leverage to perform predictions.
Our data-driven taxonomization of graph datasets provides a new understanding of critical dataset characteristics that will enable better model evaluation and the development of more specialized GNN models.
\end{abstract}
\vspace{-10pt}


\section{Introduction}

Machine learning for graph-structured data has seen rapid development in recent years \cite{hamilton2020}.
Originally inspired by convolutional neural networks, which are very successful in regular Euclidean domains thanks to their ability to leverage data-intrinsic geometries, classical graph neural network (GNN) models~\cite{defferrard2016convolutional,kipf2016GCN, velivckovic2017graph} translate those principles to graph domains.
Further advancements in the field led to a wide selection of complex and powerful GNN architectures. Some models are provably more expressive than others \cite{xu2018gin, morris2019kGNN}, can leverage multi-resolution views of graphs \cite{min_scattering_2020}, or can account for implicit symmetries of the graph data \cite{bronstein_geometric_2021}; comprehensive surveys of graph neural networks can be found in \cite{Bronstein_2017,wu2020,ZHOU202057}. However these GNN methods are historically evaluated on a set of small datasets \cite{morris2020tudataset} that became insufficient to serve as distinguishing benchmarks \cite{dwivedi_benchmarking_2020}. Therefore, recent work has focused on compiling a set of large(r) benchmarking datasets across diverse graph domains \cite{dwivedi_benchmarking_2020, hu_open_2021}. Despite these efforts and the introduction of new datasets, it is still not well-understood, for example, whether node features or (sub)graph structural patterns are more influential, or how important long-range interactions are. Hence it is not clear what aspects of GNNs' representation capabilities are tested by a given benchmark. Here, we explore graph data characteristics in relation to the prediction tasks and establish the first taxonomic view of GNN benchmarking datasets. These insights improve our understanding of empirical evaluations of GNNs and will lead to appropriate empirical validation of future models.


\section{Methods}
\vspace{-2pt}

Information in graph data is typically encoded in two ways: (i) \emph{node features} that represent the properties at individual nodes of the graph, and (ii) \emph{graph structure} that represents relations between nodes.
Leveraging symmetries and other geometric priors in graph data is crucial for generalizable learning \cite{bronstein_geometric_2021}. While invariance (i.e., symmetry) or equivariance to some transformations is inherent, invariance to others may be only empirical or partial. Here we use this lens of empirical transformation sensitivity to gauge \emph{how} task-related information is encoded in graph datasets. We perturb graph datasets with a set of transformations designed to eliminate or emphasize particular types of information embedded in the graphs. As a proxy to how invariant or sensitive a given prediction task is to these graph perturbations, we observe the empirical change in GNN performance on such perturbed versions of the datasets compared to the originals. For a given dataset and its prediction task, the empirical \emph{sensitivity profile} to the set of perturbations represents a comprehensive view of what information is important and needs to be captured by a GNN model. Based on these sensitivity profiles, we cluster the analyzed datasets and propose a taxonomy.


\begin{figure}[tb]
\centering
     \begin{subfigure}[b]{0.15\textwidth}
         \centering
         \includegraphics[width=\textwidth]{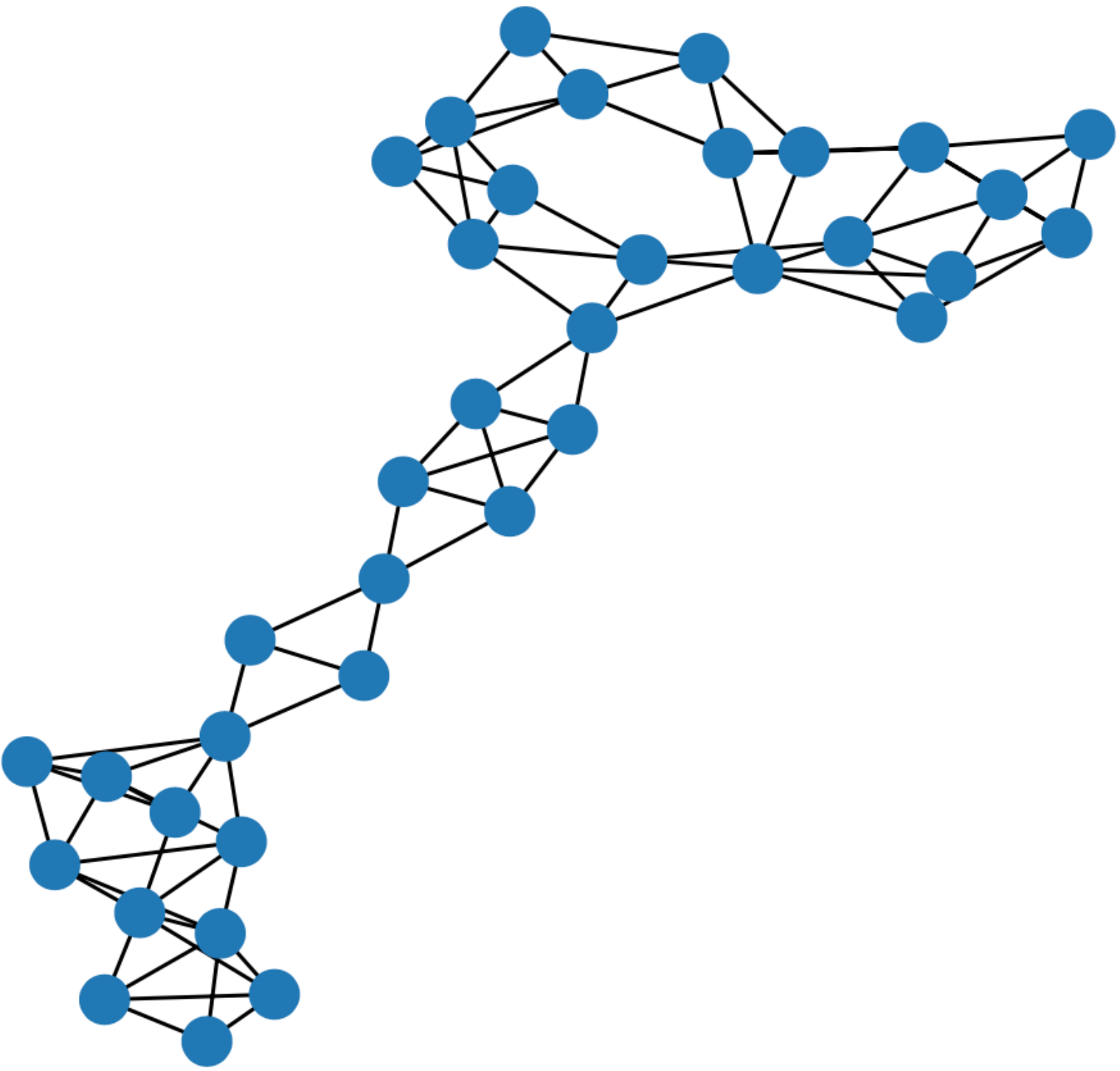}
         \caption{original}
         \label{fig:perts-og}
     \end{subfigure}
     \hfill
     \begin{subfigure}[b]{0.15\textwidth}
         \centering
         \includegraphics[width=\textwidth]{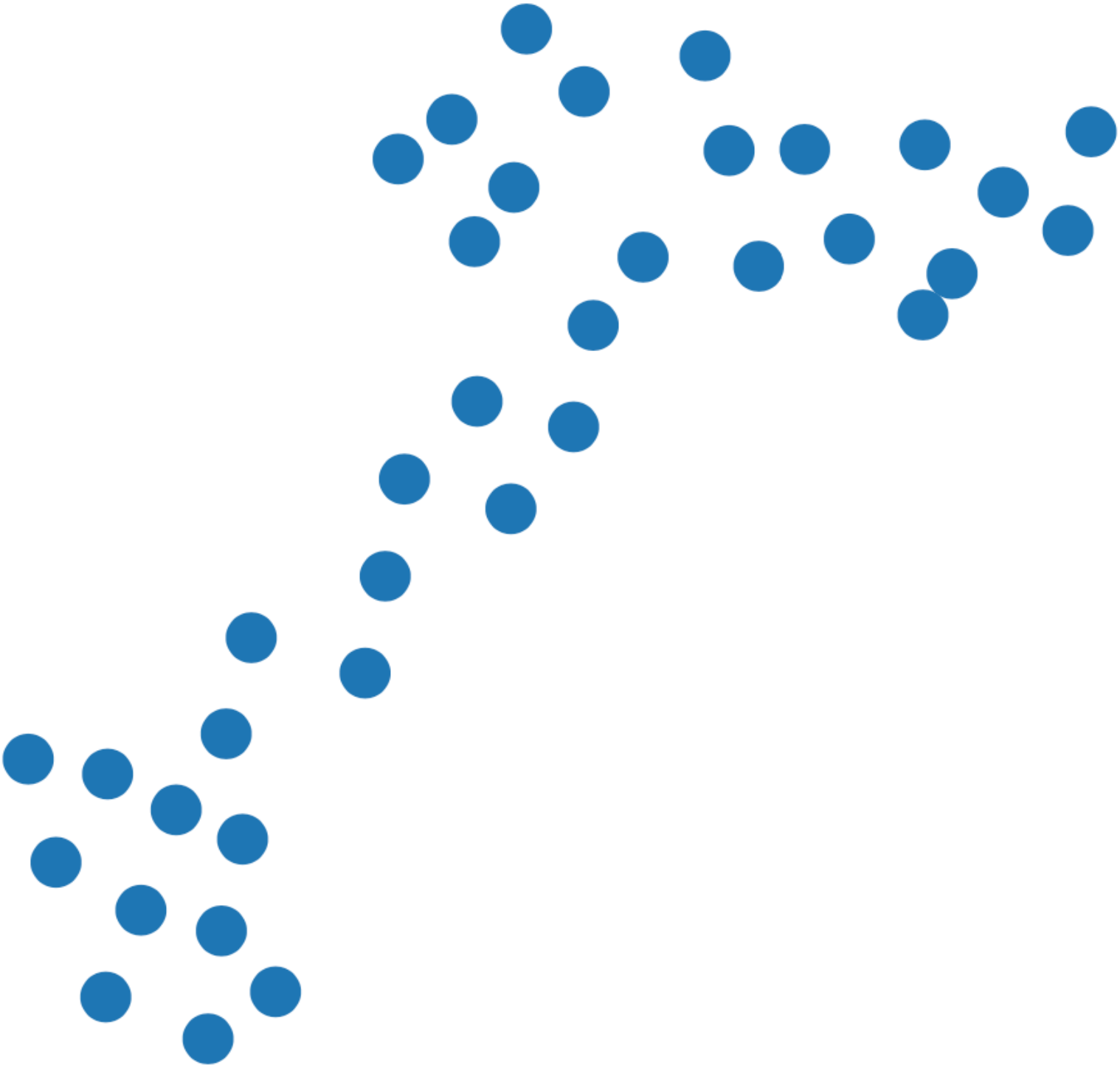}
         \caption{no-edges}
         \label{fig:perts-ne}
     \end{subfigure}
     \hfill
     \begin{subfigure}[b]{0.15\textwidth}
         \centering
         \includegraphics[width=\textwidth]{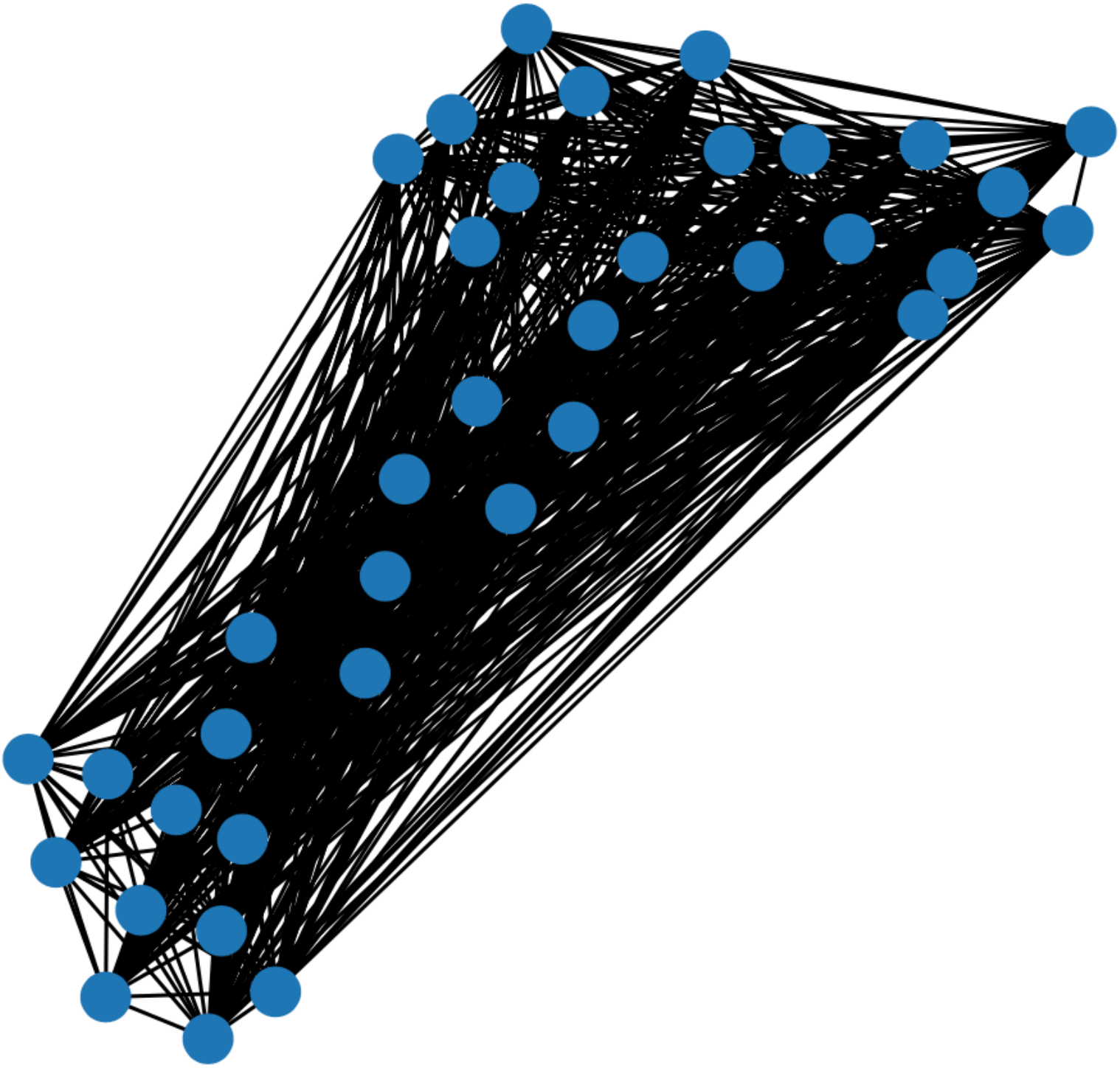}
         \caption{fully-conn.}
         \label{fig:perts-fc}
     \end{subfigure}
     \hfill
     \begin{subfigure}[b]{0.15\textwidth}
         \centering
         \includegraphics[width=\textwidth]{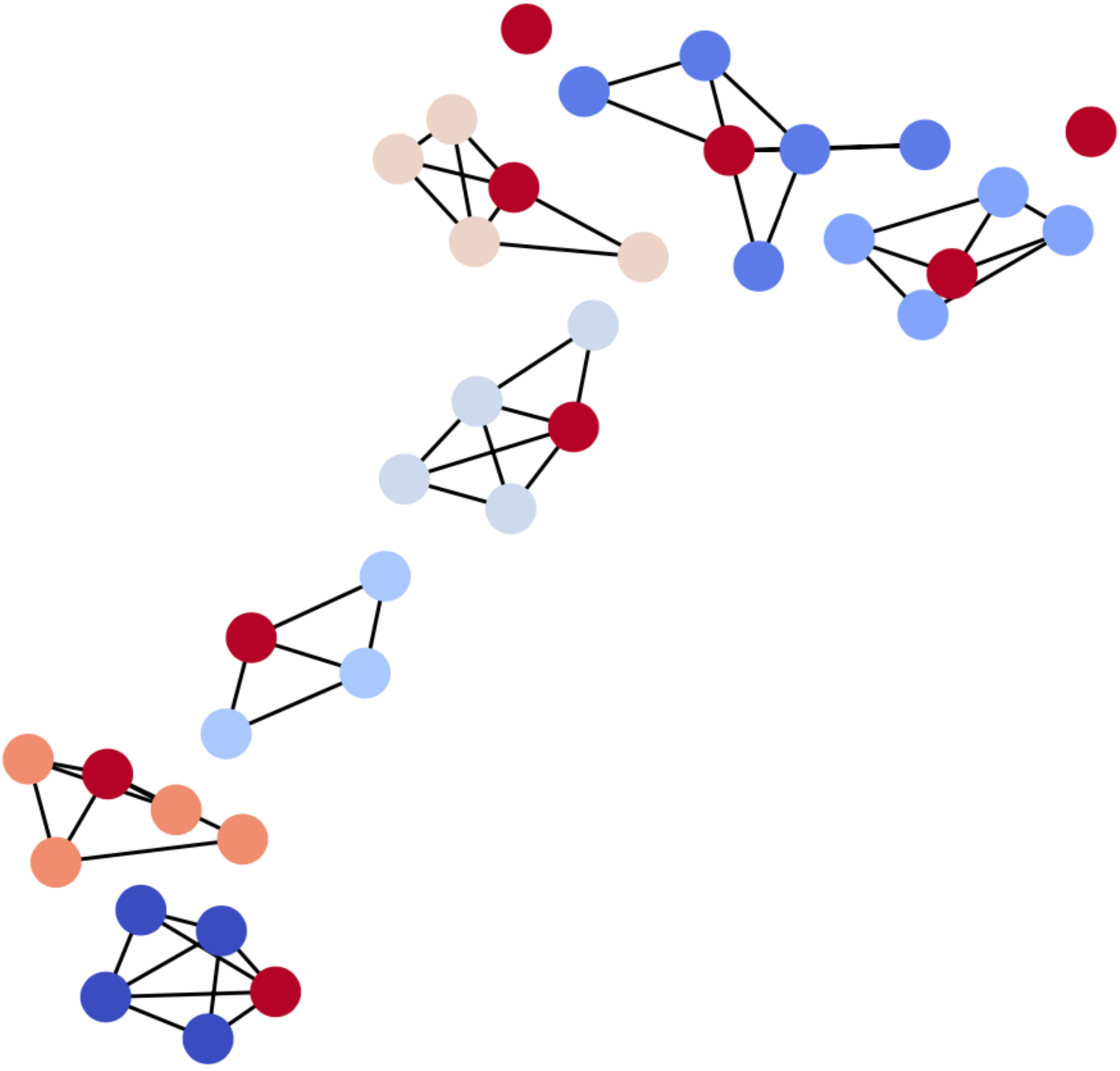}
         \caption{frag., $k=1$}
         \label{fig:perts-frag1}
     \end{subfigure}
          \begin{subfigure}[b]{0.15\textwidth}
         \centering
         \includegraphics[width=\textwidth]{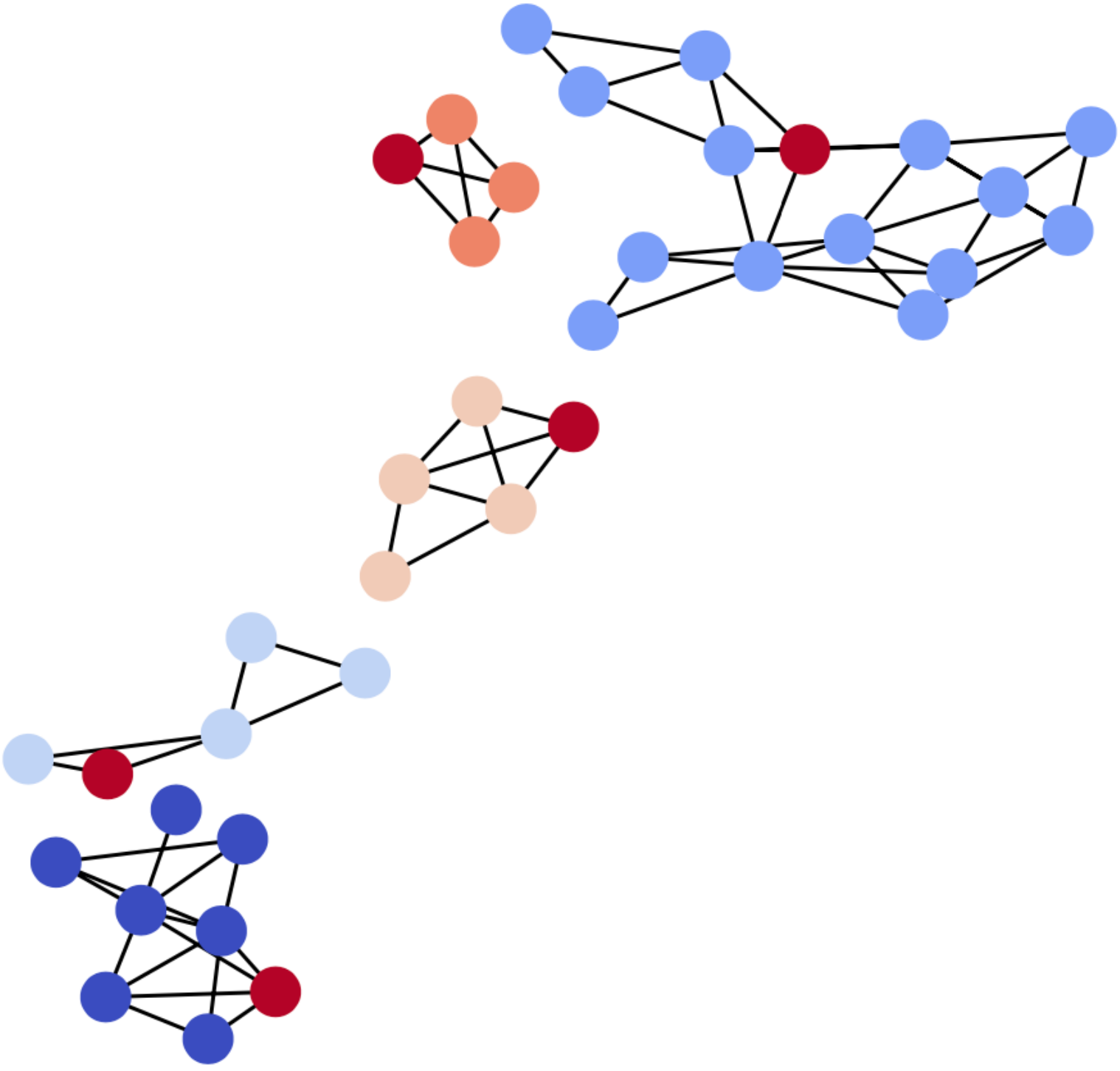}
         \caption{frag., $k=2$}
         \label{fig:perts-frag2}
     \end{subfigure}
          \begin{subfigure}[b]{0.15\textwidth}
         \centering
         \includegraphics[width=\textwidth]{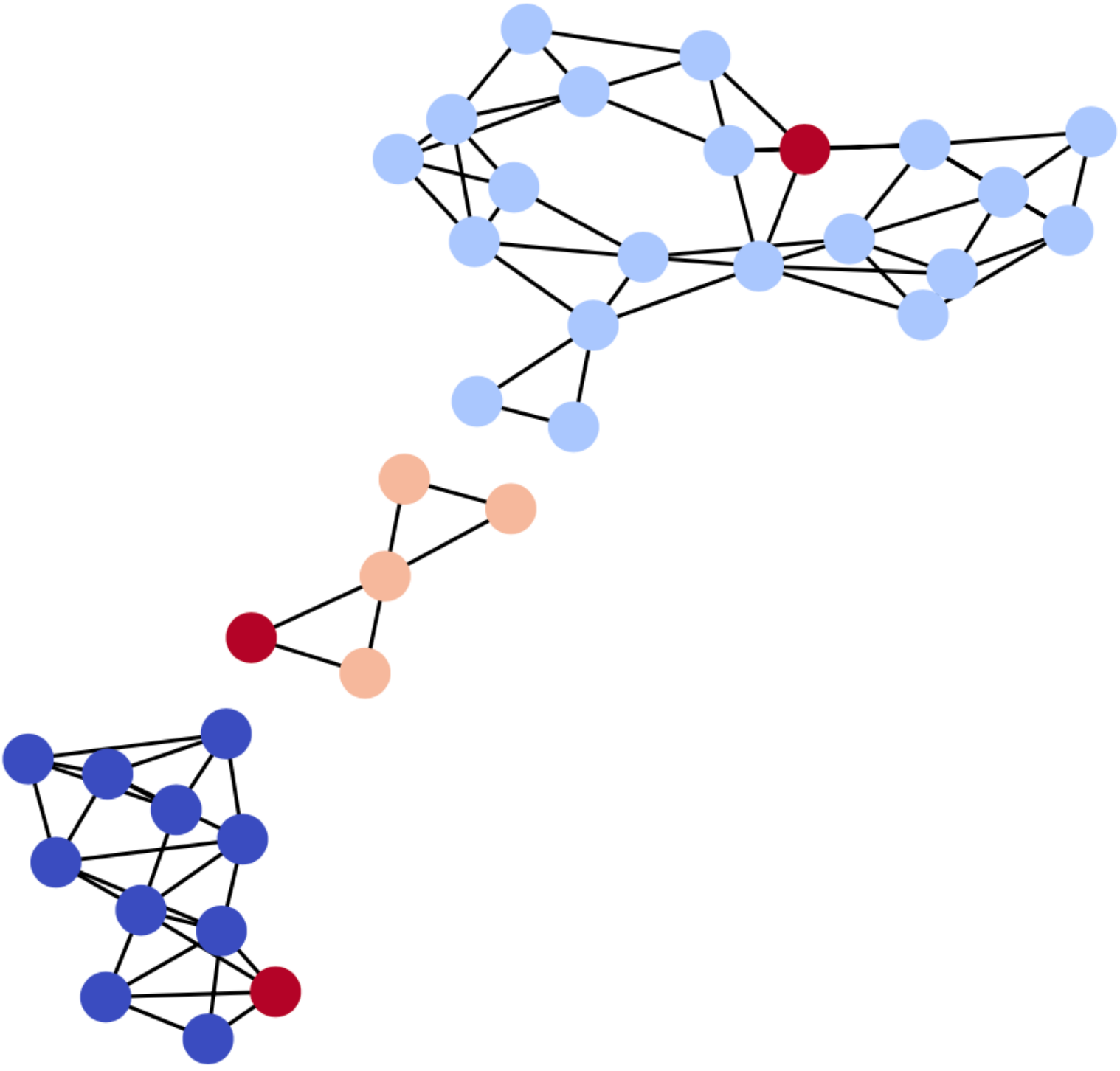}
         \caption{frag., $k=3$}
         \label{fig:perts-frag3}
     \end{subfigure}
\caption{\small \textbf{Graph structure perturbations.} An illustrative example from ENZYMES dataset.}
\label{fig:perts}
\end{figure}

First, we utilize two types of node perturbations: node features are either discarded or they are replaced by one-hot encodings of node degrees; we refer to these perturbations as \emph{no-node-features} and \emph{node-degree}, respectively. Secondly, we design a set perturbations acting on graph structure (Fig.~\ref{fig:perts}). Perturbations that remove all edges (\emph{no-edges)} or make the graph \emph{fully-connected} eliminate structural information and essentially turn the graph into a set, causing the nodes to be either processed fully independently, or collectively. Next, to inspect the importance of local vs.\ global graph structure, we design the \emph{fragmented} perturbation, which partitions the graph into connected components consisting of nodes whose distance to the seed node is less than $k$.
A smaller $k$ implies smaller components, and hence discards the global structure and long-range interactions.

\begin{figure}[tb]
\centering
     \vspace{-12pt}
     \begin{subfigure}[b]{0.49\textwidth}
         \centering
         \includegraphics[trim=0 11 10 0, clip, width=\textwidth]{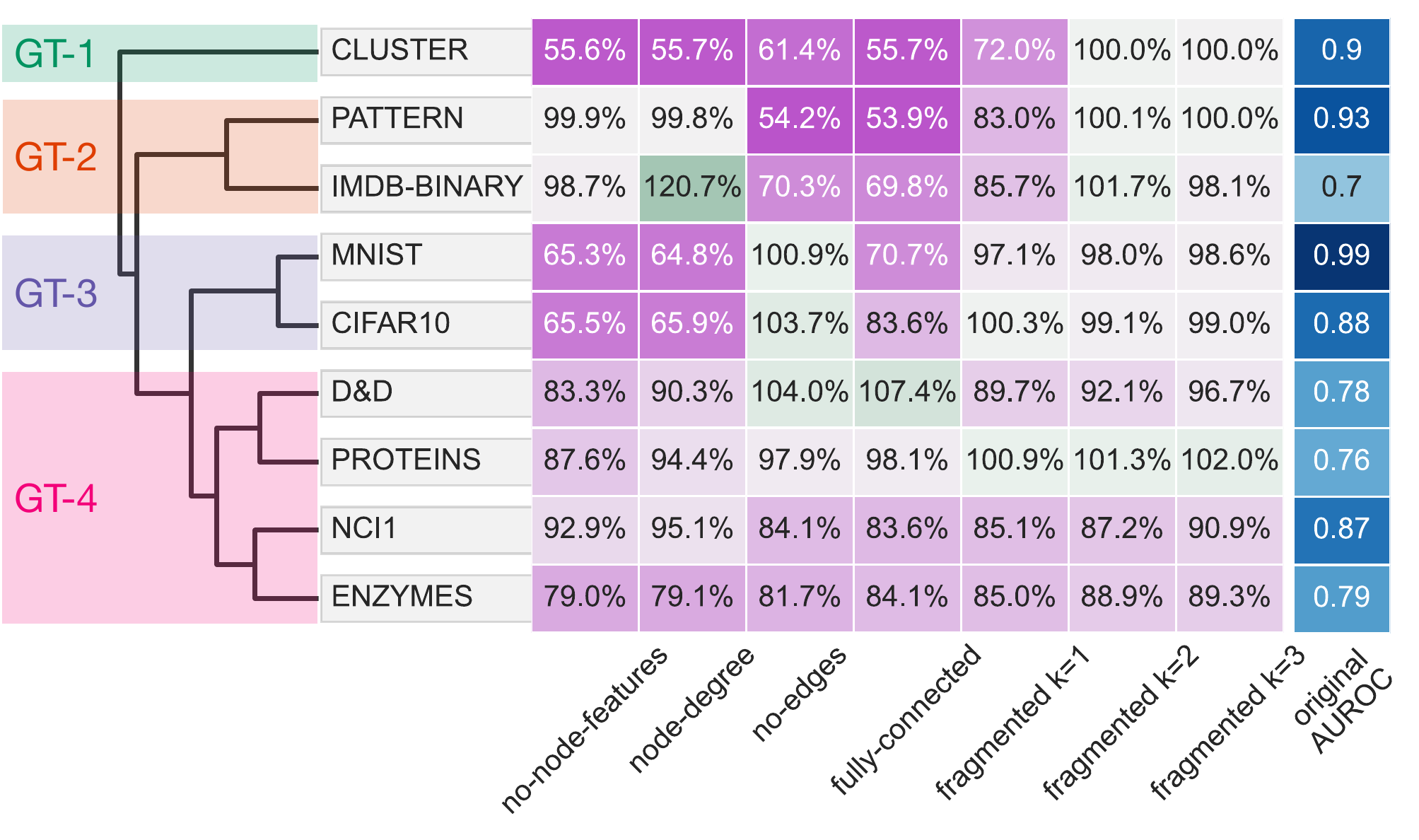}
         \caption{GCN}
         \label{fig:gl-taxo-gcn}
     \end{subfigure}
     \hfill
     \begin{subfigure}[b]{0.49\textwidth}
         \centering
         \includegraphics[trim=0 11 10 0, clip, width=\textwidth]{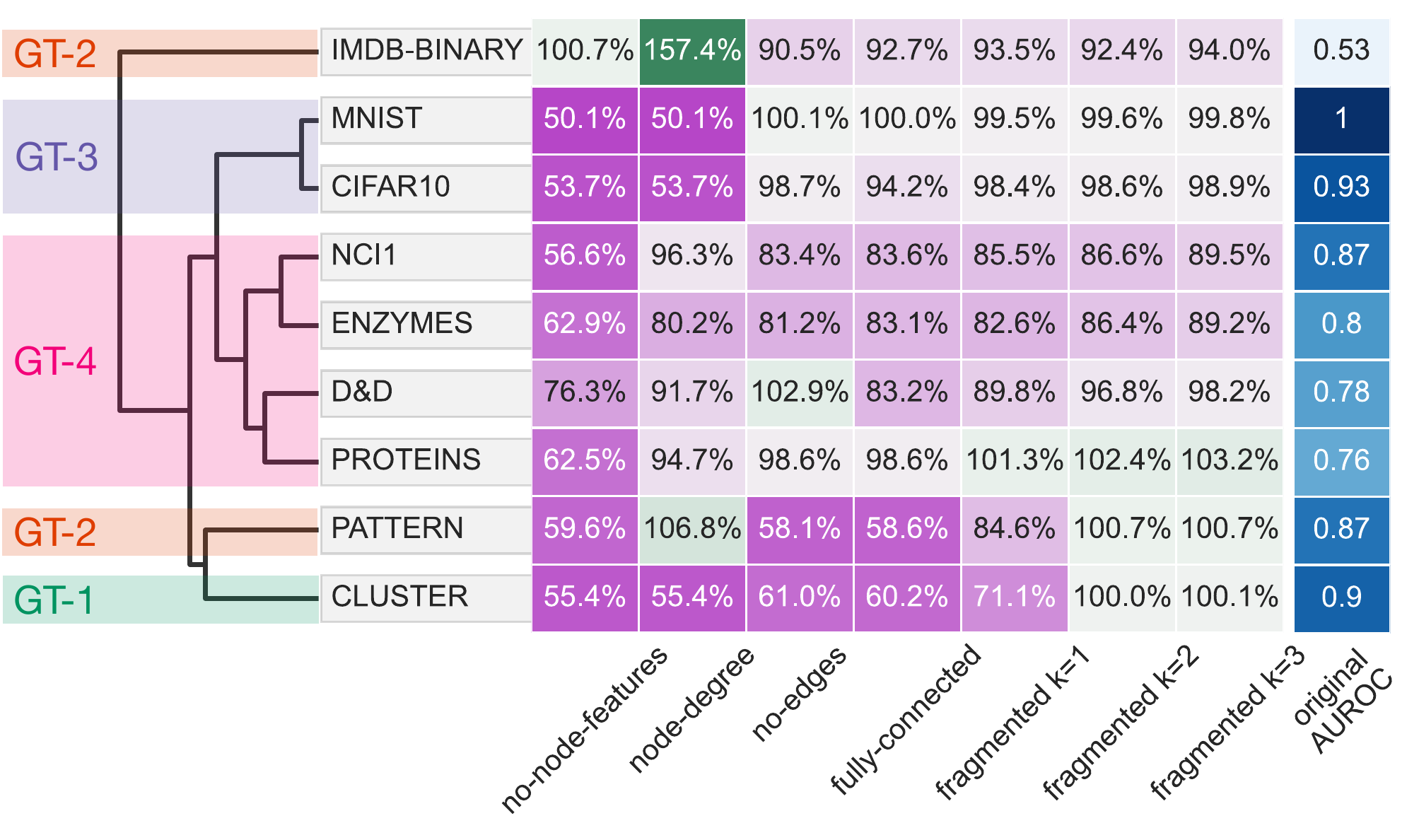}
         \caption{GAT}
         \label{fig:gl-taxo-gat}
     \end{subfigure}
     \hfill
     \begin{subfigure}[b]{0.49\textwidth}
         \centering
         \includegraphics[trim=0 11 10 -10, clip, width=\textwidth]{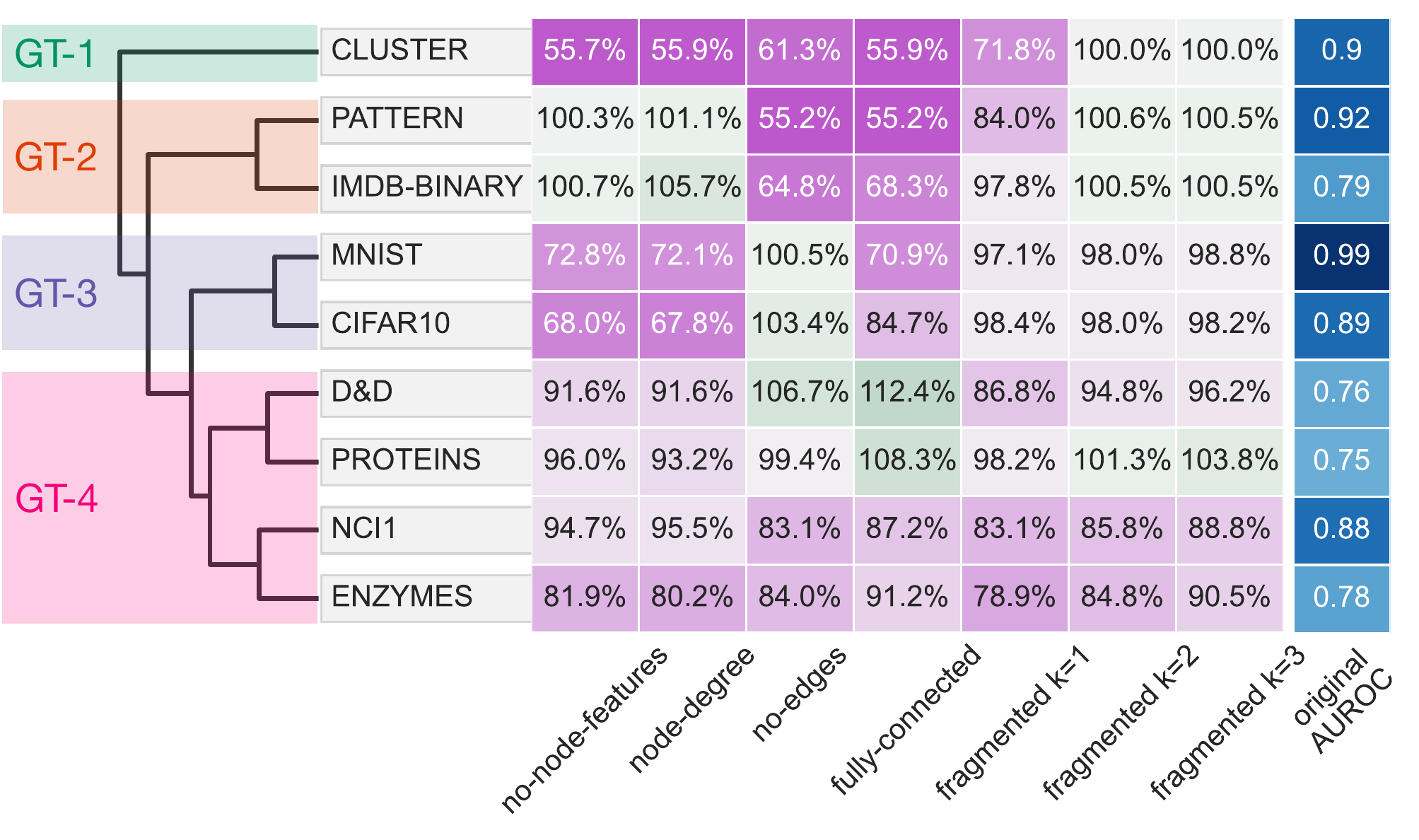}
         \caption{GIN}
         \label{fig:gl-taxo-gin}
     \end{subfigure}
     \hfill
     \begin{subfigure}[b]{0.49\textwidth}
         \centering
         \includegraphics[trim=0 11 10 -10, clip, width=\textwidth]{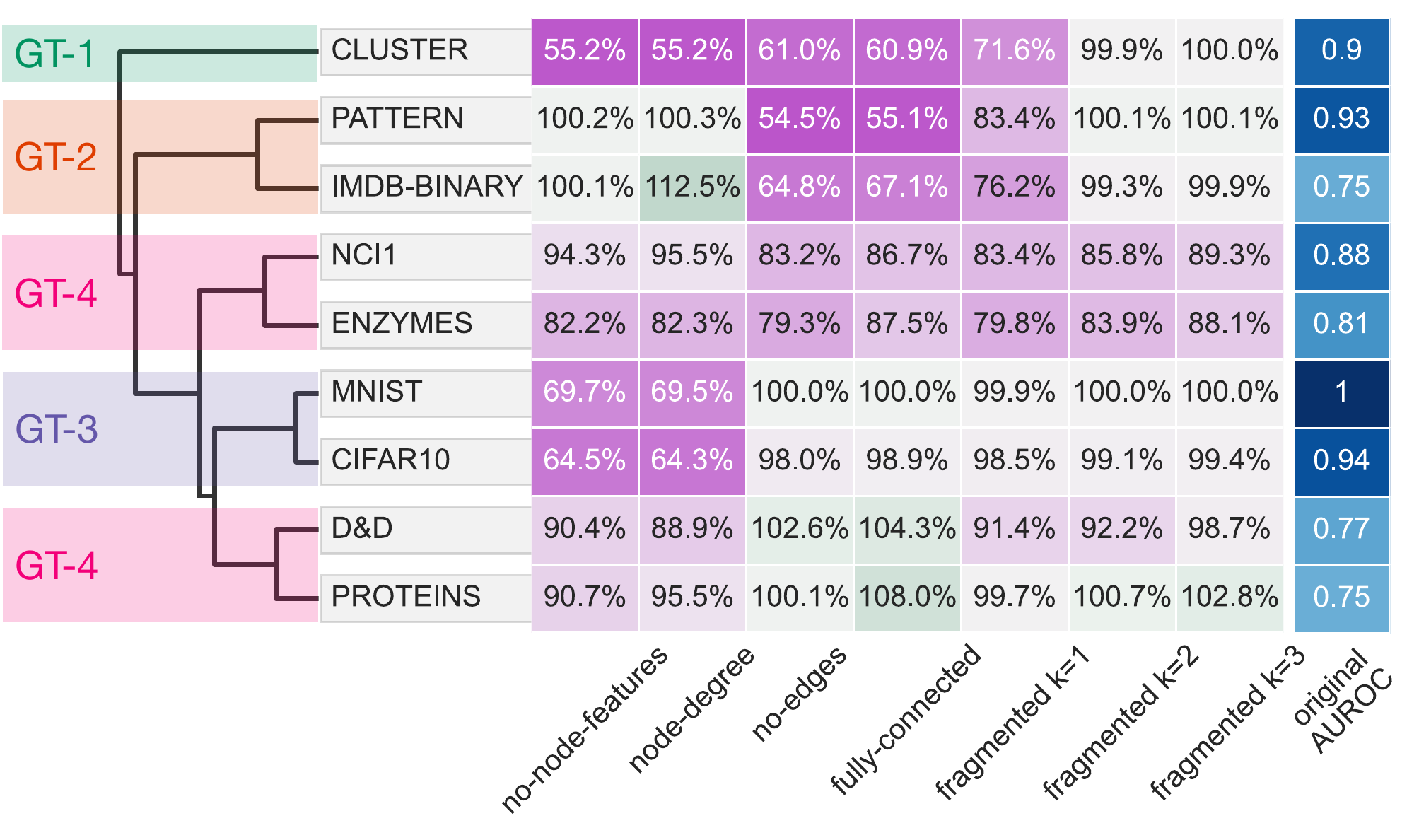}
         \caption{ChebNet}
         \label{fig:gl-taxo-cheb}
     \end{subfigure}
\vspace{-3pt}
\caption{\textbf{Taxonomy of inductive graph learning datasets via graph perturbations.} For each GNN model (a--d), dataset, and perturbation combination, we show the model's performance relative to its performance on the unmodified dataset. The categorization into 4 dataset clusters is stable across various GNN models with only minor deviation.}
\label{fig:gl-taxo}
\end{figure}

\begin{figure}[tb]
\centering
\includegraphics[trim=5 10 10 5, clip, width=0.7\textwidth]{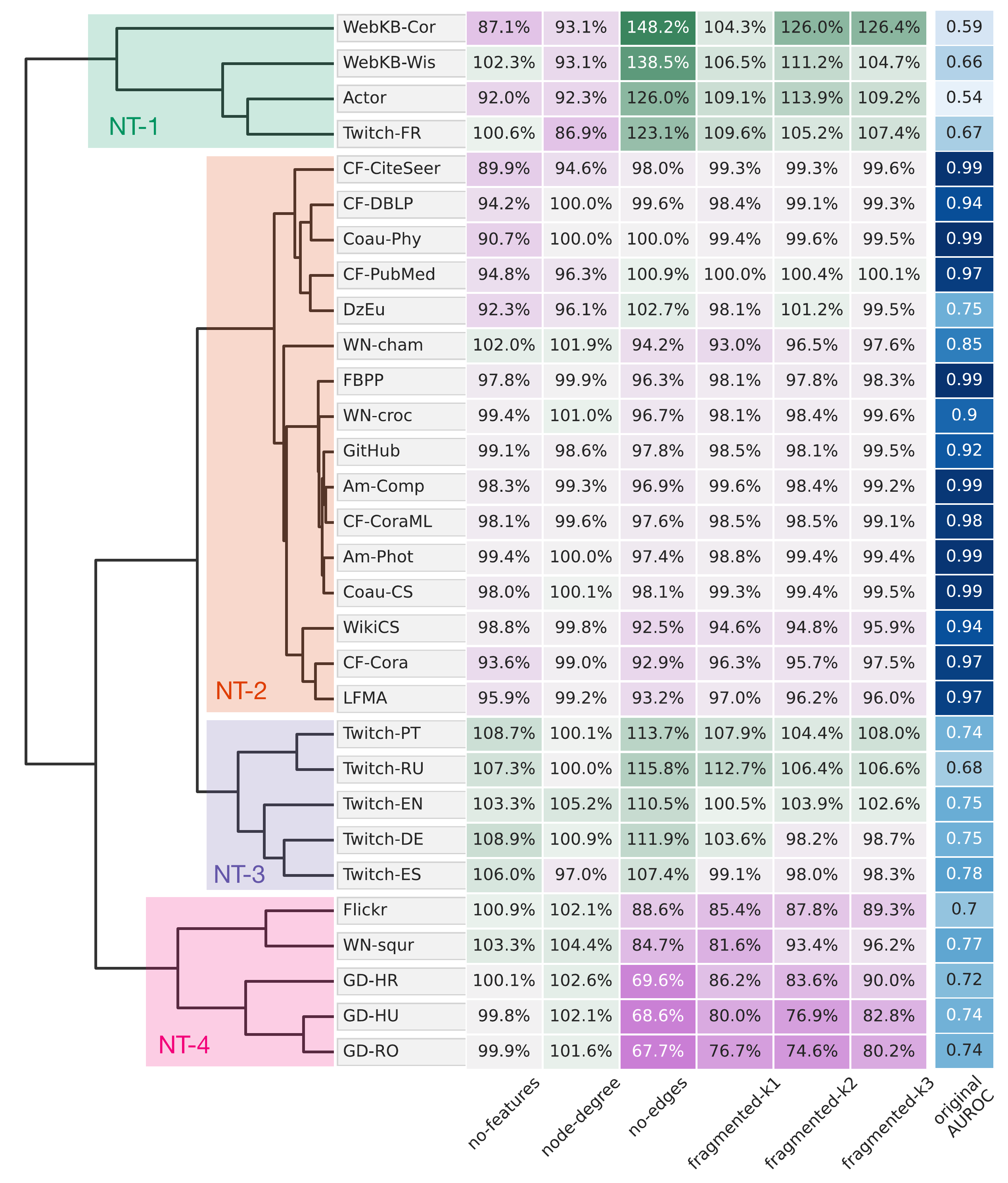}
\caption{\textbf{Taxonomy of transductive node-level prediction datasets.} Categorization into 4 dataset groups (NT-1 to NT-4) is based on clustering of sensitivity profiles w.r.t. a GCN-based model.
}
\label{fig:nl-taxo}
\end{figure}

With these perturbations, we comprehensively profile a selection of widely used graph-level classification datasets (Fig.~\ref{fig:gl-taxo}) that cover the benchmarking dataset collection of Dwivedi et al.~\cite{dwivedi_benchmarking_2020}, and a wide range of node-level classification datasets (Fig.~\ref{fig:nl-taxo}) accessed via the PyG package \cite{FeyLenssen2019PyG}.

\section{Results}

To obtain the perturbation sensitivity profiles, we consider
four popular GNN models: GCN~\cite{kipf2016GCN}, GAT~\cite{velivckovic2017graph}, GIN~\cite{xu2018gin}, and ChebNet~\cite{defferrard2016convolutional}. We keep the model hyperparameters identical for each dataset and perturbation combination: 2-layer MLP for node embedding (only for node-level tasks), 5 graph convolutional layers with residual connections and batch normalization, followed by global mean pooling (for graph-level prediction tasks), and finally a 2-layer MLP classifier. The classification performance is evaluated in terms of AUROC by averaging over either 10-fold stratified cross-validation (for datasets without a standard training/validation/test split); or averaged over 10 repetitions with different random initializations using the standard splits.



\vspace{-2pt}
\subsection{Graph-level prediction tasks}
We identify a categorization into four dataset clusters based on hierarchical clustering of their perturbation sensitivity profiles. We refer to these \textbf{g}raph-level \textbf{t}ask clusters as GT-1 to GT-4, respectively. This categorization is stable across the four GNN models with only minor deviations. Here we include CLUSTER and PATTERN datasets, that are in fact node classification datasets, but their inductive learning setup makes them more akin to graph-level tasks than to transductive node-level tasks.

Grouped in GT-3, the image-derived datasets CIFAR10 and MNIST \cite{dwivedi_benchmarking_2020} share very similar sensitivity profiles, showing that the color of underlying superpixels (node features) is the dominant information, while the graph structure is irrelevant. Interestingly, however, early global symmetry (\emph{fully-connected}) appears to lead to detrimental (over-)smoothing for GCN and GIN, while GAT and ChebNet are robust to this perturbation.

CLUSTER and PATTERN \cite{dwivedi_benchmarking_2020}, generated from a stochastic block model, as well as the actor/actress ego-network dataset IMDB-BINARY \cite{yanardag2015DGK}, are greatly affected by elimination of \emph{local} graph structure. In the latter two, which compose GT-2, the node features bear no importance: using the node degrees instead improves the performance in IMDB-BINARY by 20\%, possibly by enhancing structural information that the prediction task depends on. The profile of CLUSTER differs from the other two in that it shows reliance on both types of information; in fact, in this dataset, the node features encode class information of a few key nodes, while maintaining graph structure is essential for correct cluster prediction of the unlabelled nodes. Note that original AUROC is retained for \emph{fragmented}~$k=\{2, 3\}$, as due to the dense nature of the graph 2-hop neighborhoods for each node recover the original graph.

Finally, the (bio)molecular datasets NCI1 \cite{wale2008}, D\&D \cite{Dobson2003}, PROTEINS, and ENZYMES \cite{Borgwardt2005} are grouped into the cluster GT-4. D\&D and PROTEINS show limited or no reliance on graph structure, as opposed to NCI1 and ENZYMES.
Interestingly, even though PROTEINS and ENZYMES contain similar protein structure graphs, their different prediction tasks lead to notably different perturbation sensitivities; this illustrates the task dependency of the optimal graph representations.

\vspace{-2pt}
\subsection{Node-level prediction tasks}

For node-level prediction tasks we restrict our experiments to a GCN model as our experiments for the graph-level tasks suggest the GNN model of choice does not have a profound impact on the resulting taxonomy.
We identify four groups of node-level datasets.
First, NT-1 and NT-3 contain all the datasets from Twitch \cite{rozemberczki2021multiscale}, WebKB \cite{CRAVEN200069}, and Actor \cite{pei2020Geom-GCN}.
Both groups, particularly NT-1, benefit from removing all edges, indicating the richness of the constructed node features for the corresponding tasks without needing additional structural information.
WebKB aims to classify web pages into, e.g., student or staff.
One may imagine the web page descriptions (node features) are informative for distinguishing student and staff pages since, for example, the word ``Ph.D.'' is more likely to appear on a student page.
Meanwhile, it is less likely that student pages link between each other (edges), as suggested by the heterophile nature of the WebKB networks \cite{mostafa_local_2021}.
Notice the NT-3 datasets also benefit from removing node features instead of removing edges, as opposed to NT-1.
This difference indicates the Twitch datasets in NT-3 contain node features and structural information that are individually useful for the prediction tasks, but collectively work against each other.

NT-2 contains a broad spectrum of datasets from citation networks to social networks and web pages, which are relatively insensitive to any graph perturbation.
This implies that both the node features and the structural information are useful for the tasks; and unlike NT-3, the two sources of information are consistent with each other.
This also accounts for the relatively high AUROC scores of NT-2 datasets (averaging 0.956) compared to NT-3 (averaging 0.740).
As an example, Amazon \cite{shchur2018pitfalls} aims to classify the categories of products (nodes) given their co-purchasing relationships (edges) and the product descriptions (node features).
One may expect that both sources of information are insightful for this classification task.
Notice the citation networks (CF \cite{bojchevski2018deep}) and the streaming service social networks (DzEu, LFMA \cite{rozemberczki2020characteristic}) are slightly more dependent on the node features, indicating the node features are somewhat more informative.
Finally, NT-4 datasets are dominantly reliant on the structural information, with Flickr \cite{zeng2020graphsaint} being particularly dependent on long-range interactions, as suggested by the same performance drop caused by the edge removal and all \emph{fragmented}  perturbations.


    

\section{Conclusion}
We provide a principled approach for the taxonomization of graph datasets based on their type of prediction task signal, rather than application domain. We studied inductive and transductive classification tasks without edge features or global graph properties; for future work we plan to extend our analysis to a wider range of graph tasks, datasets, and GNN models. Our taxonomy is determined by the set of perturbations and the expressive power of the GNN models used. Combining a wider range of these views will lead to a more complete taxonomy, potentially leading to a better understanding of the modeling capabilities of GNNs as well as providing valuable insights required to improve upon existing models.



\section*{Acknowledgement}

This research was partially funded by the National Institutes of Health, NIGMS grant \#R01GM135929 [M.H., G.W], supporting R.L., S.M., M.H.; the National Science Foundation, DMS grant \#1845856 [M.H.], supporting M.H; Fin-ML scholarship [F.W.], supporting F.W.; FRQNT grant 299376 [G.W.], supporting F.W.; CIFAR AI Chair [G.W.], supporting G.W.; and IVADO grant PRF-2019-3583139727 [G.W.], supporting S.C. and L.R. The content provided here is solely the responsibility of the authors and does not necessarily represent the official views of the funding agencies.

\bibliographystyle{plain}
\bibliography{references}

\end{document}